\def\year{2021}\relax
\documentclass[letterpaper]{article} % DO NOT CHANGE THIS
\usepackage{aaai21}  % DO NOT CHANGE THIS
\usepackage{times}  % DO NOT CHANGE THIS
\usepackage{helvet} % DO NOT CHANGE THIS
\usepackage{courier}  % DO NOT CHANGE THIS
\usepackage[hyphens]{url}  % DO NOT CHANGE THIS
\usepackage{graphicx} % DO NOT CHANGE THIS
\usepackage{natbib}  % DO NOT CHANGE THIS AND DO NOT ADD ANY OPTIONS TO IT
\usepackage{caption} % DO NOT CHANGE THIS AND DO NOT ADD ANY OPTIONS TO IT
\nocopyright
\title{CLAWS: Contrastive Learning with hard Attention and Weak Supervision}
\author{
    Jansel Herrera-Gerena,\textsuperscript{\rm 1}
    Ramakrishnan Sundareswaran, \textsuperscript{\rm 1}
    John Just,\textsuperscript{\rm 1}
    Matthew Darr,\textsuperscript{\rm 1}
    Ali Jannesari\textsuperscript{\rm 1}\\
}
\affiliations{
    \textsuperscript{\rm 1}Iowa State University\\
    janselh@iastate.edu,
    ramkris@iastate.edu,
    justjo@iastate.edu,
    darr@iastate.edu,
    jannesari@iastate.edu

}
\begin{document}

\maketitle

\begin{abstract}
% Learning effective visual representations without human supervision is a long-standing problem in computer vision. Recent advances in self-supervised learning algorithms have utilized contrastive learning. One such method, SimCLR, applies a composition augmentations to an image, and minimizes a contrastive loss between the two augmented images. In this paper we present \textit{CLAWS} which combine the idea of SimCLR with supervised learning and see the effect that contrastive learning has within class clusters. In addition, we added an attention mask to a cropped image before the two images are compared with a contrastive loss function. This mask forces the network to focus on pertinent object features and ignore background features. We compare resuts betwen a supervised SimCLR and \textit{CLAWS} over an agricultural dataset with 227,060 samples of 11 different crop classes. After conducting experiments we were able acheive an NMI of 0.7325. 

Learning effective visual representations without human supervision is a long-standing problem in computer vision. Recent advances in self-supervised learning algorithms have utilized contrastive learning, with methods such as SimCLR, which applies a composition of augmentations to an image, and minimizes a contrastive loss between the two augmented images. In this paper, we present CLAWS, an annotation-efficient learning framework, addressing the problem of manually labeling large-scale agricultural datasets along with potential applications such as anomaly detection and plant growth analytics. CLAWS uses a network backbone inspired by SimCLR and weak supervision to investigate the effect of contrastive learning within class clusters. In addition, we inject a hard attention mask to the cropped input image before maximizing agreement between the image pairs using a contrastive loss function. This mask forces the network to focus on pertinent object features and ignore background features. We compare results between a supervised SimCLR and CLAWS using an agricultural dataset with 227,060 samples consisting of 11 different crop classes. Our experiments and extensive evaluations show that CLAWS  achieves a competitive NMI score of 0.7325. Furthermore, CLAWS engenders the creation of low dimensional representations of very large datasets with minimal parameter tuning and forming well-defined clusters, which lends themselves to using efficient, transparent, and highly interpretable clustering methods such as Gaussian Mixture Models.

\end{abstract}

\section{Introduction}
In the last few years, there have been several advances in deep learning and artificial intelligence for solving problems in agriculture, and a lot of this innovation is driven by a large amount of data at our disposal. More specifically, a vast amount of data is generated with information about crop fields, crop type, yield growth, plant phenotyping, and plant breeding statistics. Additionally, a lot of visual information is available that can be exploited to solve problems such as detecting anomalies; for example, taller crops showing erroneous growth and spanning larger areas can be valuable insight. While this data has been used to experiment with a wide range of machine learning and deep learning models, most of the available data in agriculture is either entirely unlabeled or partially labeled, which motivates us to tackle certain problems using an unsupervised approach. We primarily address the problems associated with the downside of data labeling, the cost of time pertaining to large-scale data labeling, and the expensive human cost and effort associated with agricultural big data. 
\\ \\
\begin{figure}[t]
  \centering
  %\fbox{\rule[-.5cm]{0cm}{4cm} \rule[-.5cm]{4cm}{0cm}}
  \includegraphics[scale=0.5]{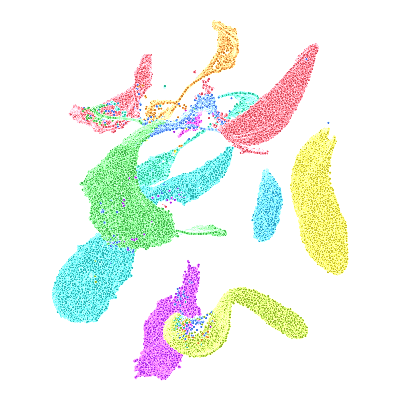}
  \caption{UMAP visualization of representations learned by CLAWS.}
\end{figure}
\begin{figure*}[t]
  \centering
  %\fbox{\rule[-.5cm]{0cm}{4cm} \rule[-.5cm]{4cm}{0cm}}
  \includegraphics[width=1\linewidth]{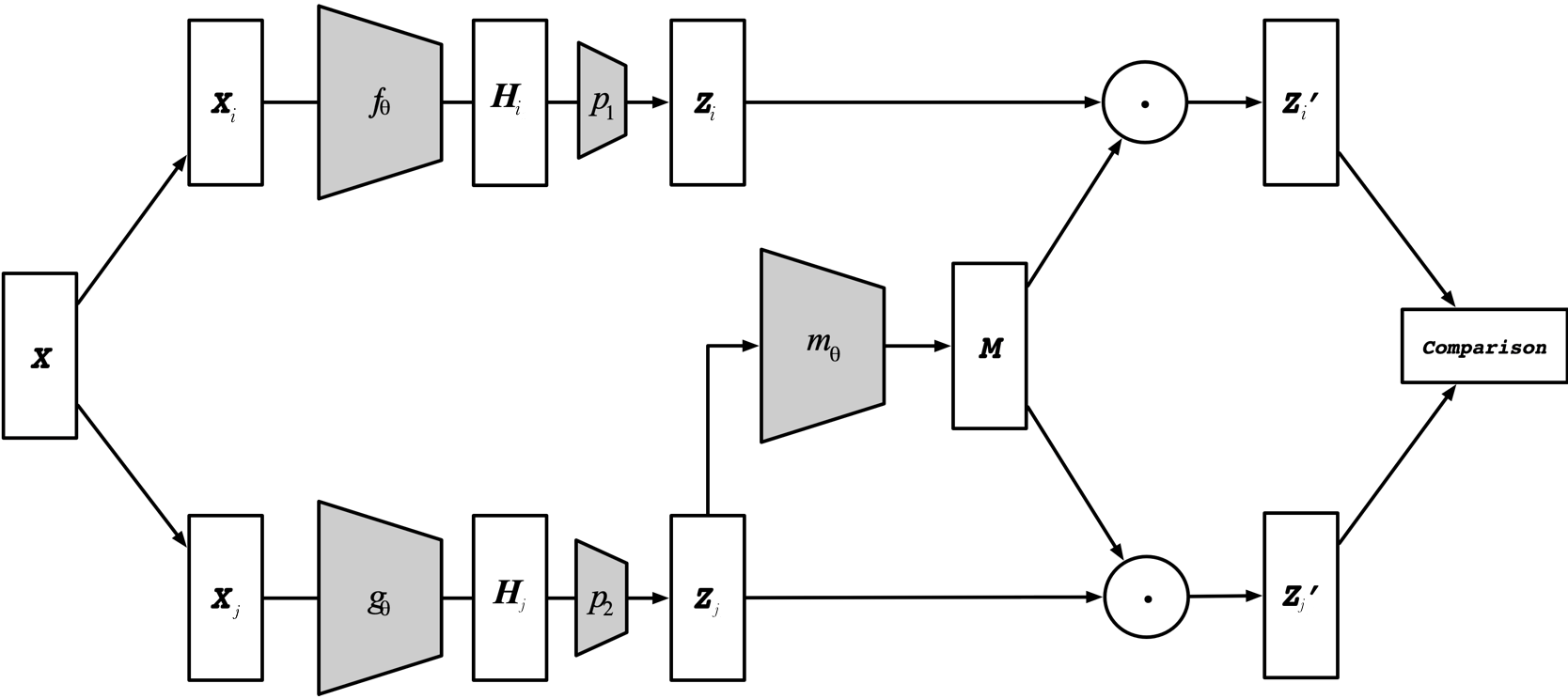}
  \caption{Illustration of the architecture in CLAWS.}
\end{figure*}
A vast majority of the modern unsupervised learning research has been driven by contrastive learning and similar self-supervised learning procedures. Contrastive learning is a popular method that learns representations of images without requiring any image labels. We directly build off of the contrastive learning framework SimCLR, presented in A Simple Framework for Contrastive  Learning of  Visual  Representations \cite{chen2020simple}. In SimCLR, each image is transformed into two correlated views. These views are then fed through a base network, ResNet, to represent each view.  These representations are then reduced in dimensionality via a Multilayer Perceptron (MLP) and compared using a contrastive loss that maximizes agreement between representations of the same image. We use the same contrastive loss as  SimCLR,  Normalized Temperature-scaled Cross-Entropy (NT-Xent). This loss is important \citep{NEURIPS2020_70feb62b} as the contrastive loss removes the notion of instance classes by directly comparing images features while the image transformation defines the invariances encoded in the features. \\ 

In this paper, we present an alteration of contrastive learning that uses \textit{focus training}. We use a method similar to \citep{chen2020simple} along with the addition of our attention head. This type of an attention head allows us to focus on important generated features for image representation. The procedure we follow consists of two networks, one to generate representations of a given input image and the second to generate representations of a \textit{crop image}. Representations from the cropped image will be passed to our attention head to focus on important features from both images representations. Lastly, the NT-Xent loss function is applied to these outputs. In addition to contrastive learning, we combine this type of learning with additional supervision to see the performance of this combination of methods while performing image clustering. In essence, we have two additions to SimCLR, the use of a crop model to generate representations of crop images and the conception of an attention model to focus the contrastive loss comparisons.

% We test the effectivity of this approach on cifar10 \citep{cifar} and examined the clusters within classes. These comparisons are based on the embeddings that each of the models produces. Results from a supervised version of the model, SimClR and \textit{CLAWS} (ours) are viewed and compared in the embedding space to examine the effectiveness of the clusters.

% \begin{figure*}[t]
% \centering
% \begin{minipage}{.5\textwidth}
%   \centering
%   \includegraphics[width=\linewidth]{embeddings_cifar_umap.png}
%   \label{fig:test1}
% \end{minipage}%
% \begin{minipage}{.5\textwidth}
%   \centering
%   \includegraphics[width=\linewidth]{embeddings_simclr_umap.png}
%   \label{fig:test2}
% \end{minipage}
%  \caption{Umap of the embeddings generated with \textit{CLAWS} (left) and SimCLR (right). }

% \end{figure*}

At a higher level, contrastive learning works by performing deep comparisons on images and backpropagating errors on the framework used to do these image representations. Before we were able to do this type of training, we relied on methods like principal component analysis \citep{ZITKO1994718}, independent component analysis \citep{ica} and self-organizing maps \citep{KOHONEN19981} between other methods that were traditional and more classical but were the key to open the idea of what we have today as deep representation learning methods. While these classical methods help lay a good foundation, there has been a lot of progress in the last few years, and we have numerous methods now that extend the idea behind contrastive learning in different ways. We can start with SimCLR, where the creation of augmented images enforces the training process; Big Self-Supervised Models are Strong Semi-Supervised Learners (SimCLRv2) \citep{chen2020big} in which they work with the combination of label amounts and network sizes. SwAV \citep{caron2021unsupervised} takes contrastive learning methodology without having pairwise comparisons. Contrastive Clustering \citep{li2020contrastive} introduces a cluster-level contrastive head in combination with an instance-level head, and Cluster Analysis with Deep Embeddings and Contrastive Learning \citep{sundareswaran2021cluster} uses a three-prolonged approach with an instance-wise contrastive head, a clustering head, and an anchor head to perform efficient image clustering. Although our idea is based on these preceding works, other interesting works (\citep{bachman2019learning, oord2019representation, wu2018unsupervised, hjelm2018learning, he2020momentum,  DBLP:journals/corr/abs-2006-07733, Chen2021ExploringSS}) have also been proposed that use contrastive learning and bring out different perspectives. 

% When talking about contrastive learning we have to also address the loss functions. Much of these loss functions rely on the comparisons between positive pairs and negative pairs, i.e, similar pairs and dissimilar pairs. For this work, we focused on the loss used in SimCLR, NT-Xent which compares samples within the same batch. \citep{Schroff_2015} consist of triplets containing two positive pairs and one negative pair, these triplets separate the positive from the negative by distance margin as they mentioned, similar to this idea \citep{1467314}. Losses that take into consideration different points of view and timesteps as Time Contrastive Networks \citep{sermanet2018timecontrastive} and CPC \citep{oord2019representation} in which they introduce the latent space to capture information useful for future samples. 

\begin{figure*}[t!]
\centering
\begin{minipage}{.5\textwidth}
  \centering
  \includegraphics[width=\linewidth]{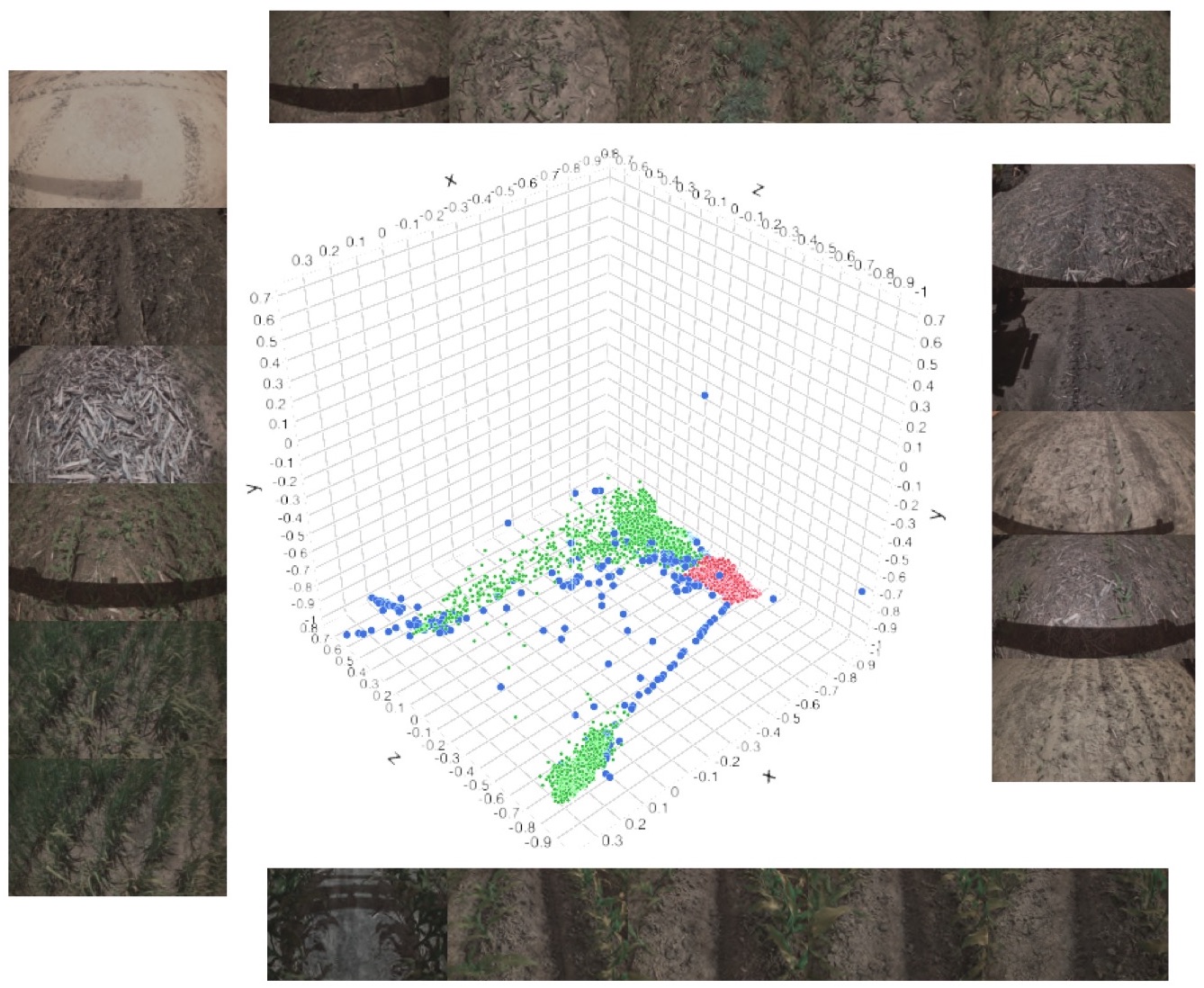}
  \label{fig:test1}
\end{minipage}%
\begin{minipage}{.5\textwidth}
  \centering
  \includegraphics[width=\linewidth]{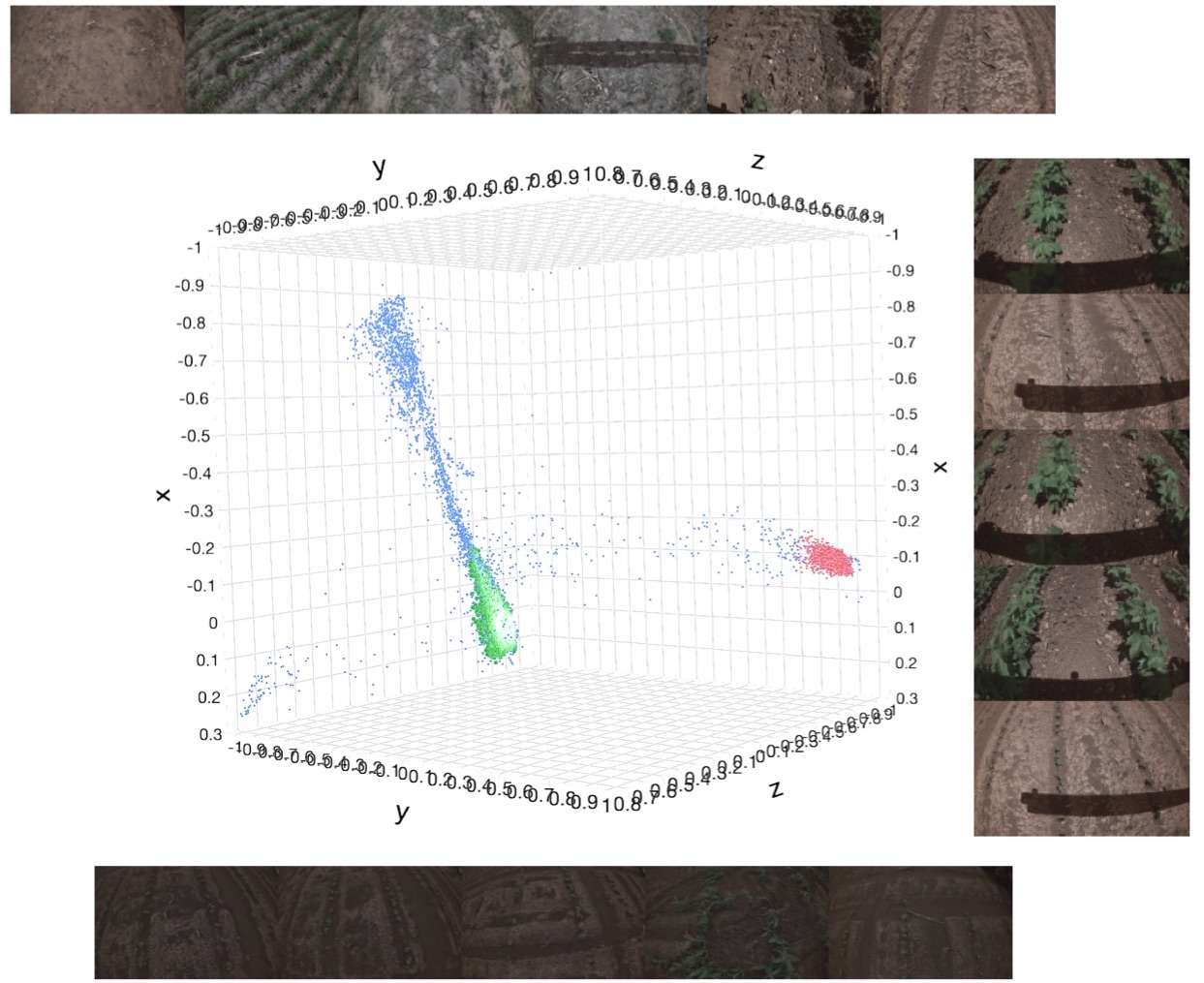}
  \label{fig:test2}
\end{minipage}
 \caption{3D plot containing clusters and outliers (blue) within a specific class. \textit{Left} represents Corn samples and \textit{Right} is showing Cotton samples. For \textit{Left} top and bottom row represents green cluster, images at the right represents red cluster and left represents blue cluster. The \textit{Right} has a column of images at the right that represents red cluster, bottom images are to show the green cluster and finally the top images represents outliers. }
\end{figure*}

\section{Methods}

An overview of our model architecture is given in Figure 2.  Similar to SimCLR, our method learns representations by maximizing agreement between differently augmented views of the same image. The first augmentation retains the full image size and is fed into a base network that is optimized for full-size images. The second augmentation takes a small crop from the original image and is fed into a base network that is optimized for small crops of images. An attention mask is then applied to both of the representations generated by the base networks,  and the attention-filtered representations are then compared using a contrastive loss. The differences between our model and SimCLR can be categorized into 3 major components: 1) the way images are fed into for inference before comparison,  2) the use of two separate base networks instead of one siamese network, and 3) application of an attention mask to the output of each base network before the contrastive loss is calculated. This is also complemented with the use of weak supervision, the two parts of our model (full image network and crop image network) are trained by using cross-entropy loss and NT-Xent.

Our models outputs can be seen as:
\begin{equation}
\textit{\textbf{H}}_i=f_{\theta}(\textit{\textbf{X}}_i)=\textnormal{EfficientNet}(\textit{\textbf{X}}_i)
\end{equation}
\begin{equation}
\textit{\textbf{H}}_j=g_{\theta}(\textit{\textbf{X}}_j)=\textnormal{EfficientNet}(\textit{\textbf{X}}_j)
\end{equation}

were $\textnormal{\textbf{X}}_i$ and $\textnormal{\textbf{X}}_j$ are a pair of augmented images drawn from \textbf{X} which is the original image. The hidden layers are $\textit{\textbf{H}}_i$ and $\textit{\textbf{H}}_j$ representing the output of the $f_{\theta}$ and $g_{\theta}$, now the hidden layers are then passed to a projection head to produce image representation with 32 features, i.e, $\textnormal{\textbf{Z}}_i=p_1(\textit{\textbf{H}}_i)$ and $\textnormal{\textbf{Z}}_i=p_2(\textit{\textbf{H}}_j)$. To calculate the attention output we send $\textnormal{\textbf{Z}}_i$ to our attention network $m_{\theta}$ to generate \textbf{\textit{M}} which is a mask that use to focus training on the important features (further explanation in section 3.3). 

\subsection{Dataset}
The single dataset used for this research is composed of 11 classes of images. Data was obtained while driving over different fields containing one of the possible crops types. A total of 227,060 samples were collected and labeled. The time in which they were collected varies, therefore there is a different variation of lighting over the dataset. The labels of the data are composed by crop fields of Wheat, Cotton, Sorghum, Corn, Peanuts, No crop, Oats, Soybeans, Canola, Sugar, WheatStubble.

\subsection{Preprocessing}

SimCLR augmented images with a composition of a random crop, random flips (horizontal and vertical), and color distortion.   This composition is applied twice to the same image to generate two distinctly augmented images. We follow the same augmentation process, except performing Gaussian blur.  We apply two separate augmentation compositions to generate augmented images (full image and crop image). Note that when SimCLR applied random cropping, the crops were always resized to 100x100 to be fed into the base network.  Our base networks have different input sizes, so we do not resize images after they are cropped. 

We use two distinct base networks, $f_{\theta}$ and $g_{\theta}$ for the full and crop networks respectively. The network used for the full image takes RGB images of size 120x190 as input, while the crop network takes RGB images of size 32x32 as input.  Both networks are a standard EfficientNet \citep{tan2020efficientnet} architecture,  and both produce output embeddings of the same size, 32 features. Using two separate base networks,  rather than feeding both image augmentations through the same base network as SimCLR does, provides two noticeable advantages.  First, we do not need to enlarge small crops to fit an architecture, which can produce unintended artifacts in the image.  Second, each architecture is more specifically optimized for global or local views.

\subsection{Attention Mask}
The main innovation in our framework is the incorporation of an attention mask.  The mask is generated with a 2-layer perceptron that takes the output of $g_{\theta}$ as input.  The output tensors of both $g_{\theta}$ and $f_\theta$ are multiplied by this mask before they are compared using a contrastive loss. This created the attention mechanism that we wanted, we created hard attention meaning that the output from the attention head is going to be either 0 or 1 (\textbf{\textit{M}}). Therefore ${\textbf{Z}^{\prime}}_i = {\textbf{Z}}_i \cdot {\textbf{\textit{M}}}$ and ${\textbf{Z}^{\prime}}_j = {\textbf{Z}}_j \cdot {\textbf{\textit{M}}}$ to then be use to calculate the loss using NT-Xent.

% The mask allows information that is not shared between the full image embedding and the cropped image embedding to be filtered out.  The mask network learns to highlight regions of the embeddings that both representations typically have in common.  In other words, the mask network learns which regions of the embeddings it should be “paying attention to” in order to maximize agreement between the values that are not filtered out. This forces $f_\theta$ to distinctly learn each component of an image that could be represented in $g_\theta$.  Consider an example where the full image is a dog standing in a grassy field, and the cropped image contains mostly just the dog’s tail. The full model needs to distinctly encode a “dog ear” concept, a “this is a tail” concept, and a “grass” concept, since the cropped image may only include one of these. $f_\theta$ only encoding a general “this is a dog” concept would not be sufficient, since it would be very difficult for $g_{\theta}$ to distinguish a dog’s tail from other animals’ tails and encode a “this is a dog” concept. It is much easier for $g_{\theta}$ to simply detect a tail.

Both base networks need to be able to encode the same concepts in the same corresponding regions of the embeddings they produce.  The same mask is applied to both embeddings, so if the output of $g_{\theta}$ contains mainly a strong concept, the mask network will remove out every region except that. If the same region in $f_\theta$ output contains a similar concept, then the two embeddings will be very similar after the mask is applied. 

% \subsection{Supervision}
% For enforcing the supervision we use the output of $p_1$ and $p_2$ and send it to two 2-layer MLP (one for $p_1$ and another for $p_2$). These MLPs allow us to predict and calculate a loss for class labels, then send them to a cross-entropy loss function. Consequently, the model is adjusted using NT-Xent and cross-entropy (for full image and crop image sides). 

\section{Results}
The models were trained for 300 epochs with a batch size of 55 and an Adam optimizer. We train with this batch size because of the way we were gathering the images for the step, here instead of randomly picking images we specify 5 images per class so that in every step we can train the model in each class.

\subsection{Evaluation}
To perform evaluation we wanted to focus on the quality of the image representation. Therefore, we compared ours against SimCLR with an adaptation, similar to our model we added a supervised section for it. Now, SimCLR and CLAWS were trained using NT-Xent in combination with supervision by adding an MLP to generate labels from the outputs of the projection head. This way we can do a fair comparison over our agricultural dataset. To evaluate both models after training we did not used the classifier section, we generated the representation of the images, i.e., ran the models over each image up to the output of the projection heads. Given the creation of all the represenation, we pass it to a K-Means clustering method and generated labels for each data point. Finally, we calculated the quality by computing NMI, AMI and ARI scores. 

\begin{table}[h!]
\centering
 \begin{tabular}{|c | c | c | c|} 
 \hline
 Method & NMI & ARI & AMI \\ [0.5ex] 
 \hline\hline
 SimCLR & 0.6101 & 0.2873 & 0.6101 \\ 
 \textit{CLAWS}(Ours) & 0.7325 & 0.5069 & 0.7324 \\ [1ex] 
 \hline
 \end{tabular}

\end{table}

In table 1 we can see that our results outperforms SimCLR with supervision. This is also specific for the quality of the represenation of the images. Meaning that by focusing the training with the use of an attention head does helps on focusing on features that relate between the corp image and the full image. 

% \begin{figure}[h!]
% \centering
%   \includegraphics[width=\linewidth]{sc_good_attn}
%   \label{fig:test1}
%  \caption{Sprite image generated by taking image representation and placing similar ones close to each other.}
% \end{figure}

% \begin{figure}[h!]
% \centering
%   \includegraphics[width=\linewidth]{attn_embeddings_tsne}
%   \label{fig:test1}
%  \caption{TSNE.}

% \end{figure}

\subsection{Gaussian Mixture Models}

Furthermore, we implemented Gaussian Mixture Models (GMM) over our image representation. Now we did not used the full dataset, we focus on seeing how our model behaved within a class. To perform this we picked \textit{Corn} and \textit{Cotton} and set the output of GMM to be two clusters and an additional one to detect outliers. In figure 3, we can see the output of GMM in the two mentioned crop types. As mentioned before  the \textit{Left} represents Corn and the \textit{Right} represents Cotton. For \textit{Left} image the top and bottom row represents green cluster, images at the right represent red cluster and the left represents blue (outliers) cluster. The \textit{Right} has a column of images at the right that represents red cluster, the bottom images are to show the green cluster, and finally, the top images represent outliers. An important thing to mention about figure 3 is that we are randomly choosing three dimensions from the 32 that represent an image. Therefore, we are not going to see the full relation between clusters in just three dimensions but we can see that even with the use of the raw representation there is a relation preserved in the plot. 

Focusing on outliers is our main reason for using GMM, having the ability to generate image representation on then pass it to these models to detect outliers can help in the reduction of the inspection time. It also allows fast customization depending on the task, because it is an unsupervised method, we can change the cluster dimensions and get fast results from it. An example of its use can be seen in figure 3, \textit{Left} contains 19,792 samples of Corn the GMM detects 315 outliers in which if we inspect them, images that have residue on the ground, are empty, the crop wrongly planted or bad image capture. This means that even when the model generates a similar representation for a specific class, we can detect, within a class, bad samples. Another example, \textit{Right} shows samples from Cotton, with 66,053 samples and GMM detected 3,401 outliers in which we can also see that for both classes the outliers resemble. Here, again, the images were empty, not correctly taken, contain residues and bad plantation. In addition to GMM clusters, if we talk about the other clusters, we can notice that there is no big issue separating them. The reason to be separated is that these crops can be going through a different stage of growth or other reasons. In figure 3, \textit{Left} top row we see images that contain weeds, the bottom row show images captured closer.  On the \textit{Right} bottom, we notice images with tire marks on the ground. Therefore, they are assigned to another cluster (within a class) because of many reasons (leaf size, plant height, and image brightness between other factors).

\section{Conclusion}
This work built upon SimCLR to achieve better representations of images using contrastive learning combined with supervision. Our framework incorporated two distinct base networks and an attention mask, which allowed the network to learn and recognize parts that strongly represent images. With this methodology, we created CLAWS and were able to show encouraging results within class clusters. Further work on this relies on completely unsupervised training.\\

\bibliography{references} 

\begin{thebibliography}{17}
\providecommand{\natexlab}[1]{#1}
\providecommand{\url}[1]{\texttt{#1}}
\providecommand{\urlprefix}{URL }
\expandafter\ifx\csname urlstyle\endcsname\relax
  \providecommand{\doi}[1]{doi:\discretionary{}{}{}#1}\else
  \providecommand{\doi}{doi:\discretionary{}{}{}\begingroup
  \urlstyle{rm}\Url}\fi

\bibitem[{Bachman, Hjelm, and Buchwalter(2019)}]{bachman2019learning}
Bachman, P.; Hjelm, R.~D.; and Buchwalter, W. 2019.
\newblock Learning Representations by Maximizing Mutual Information Across
  Views.

\bibitem[{Bell and Sejnowski(1995)}]{ica}
Bell, A.~J.; and Sejnowski, T.~J. 1995.
\newblock An Information-Maximization Approach to Blind Separation and Blind
  Deconvolution.
\newblock \emph{Neural Comput.} 7(6): 1129–1159.
\newblock ISSN 0899-7667.
\newblock \doi{10.1162/neco.1995.7.6.1129}.
\newblock \urlprefix\url{https://doi.org/10.1162/neco.1995.7.6.1129}.

\bibitem[{Caron et~al.(2020)Caron, Misra, Mairal, Goyal, Bojanowski, and
  Joulin}]{NEURIPS2020_70feb62b}
Caron, M.; Misra, I.; Mairal, J.; Goyal, P.; Bojanowski, P.; and Joulin, A.
  2020.
\newblock Unsupervised Learning of Visual Features by Contrasting Cluster
  Assignments.
\newblock In Larochelle, H.; Ranzato, M.; Hadsell, R.; Balcan, M.~F.; and Lin,
  H., eds., \emph{Advances in Neural Information Processing Systems},
  volume~33, 9912--9924. Curran Associates, Inc.
\newblock
  \urlprefix\url{https://proceedings.neurips.cc/paper/2020/file/70feb62b69f16e0238f741fab228fec2-Paper.pdf}.

\bibitem[{Caron et~al.(2021)Caron, Misra, Mairal, Goyal, Bojanowski, and
  Joulin}]{caron2021unsupervised}
Caron, M.; Misra, I.; Mairal, J.; Goyal, P.; Bojanowski, P.; and Joulin, A.
  2021.
\newblock Unsupervised Learning of Visual Features by Contrasting Cluster
  Assignments.

\bibitem[{Chen et~al.(2020{\natexlab{a}})Chen, Kornblith, Norouzi, and
  Hinton}]{chen2020simple}
Chen, T.; Kornblith, S.; Norouzi, M.; and Hinton, G. 2020{\natexlab{a}}.
\newblock A Simple Framework for Contrastive Learning of Visual
  Representations.

\bibitem[{Chen et~al.(2020{\natexlab{b}})Chen, Kornblith, Swersky, Norouzi, and
  Hinton}]{chen2020big}
Chen, T.; Kornblith, S.; Swersky, K.; Norouzi, M.; and Hinton, G.
  2020{\natexlab{b}}.
\newblock Big Self-Supervised Models are Strong Semi-Supervised Learners.
\newblock \emph{arXiv preprint arXiv:2006.10029} .

\bibitem[{Chen and He(2021)}]{Chen2021ExploringSS}
Chen, X.; and He, K. 2021.
\newblock Exploring Simple Siamese Representation Learning.
\newblock In \emph{CVPR}.

\bibitem[{Grill et~al.(2020)Grill, Strub, Altch{\'{e}}, Tallec, Richemond,
  Buchatskaya, Doersch, Pires, Guo, Azar, Piot, Kavukcuoglu, Munos, and
  Valko}]{DBLP:journals/corr/abs-2006-07733}
Grill, J.; Strub, F.; Altch{\'{e}}, F.; Tallec, C.; Richemond, P.~H.;
  Buchatskaya, E.; Doersch, C.; Pires, B.~{\'{A}}.; Guo, Z.~D.; Azar, M.~G.;
  Piot, B.; Kavukcuoglu, K.; Munos, R.; and Valko, M. 2020.
\newblock Bootstrap Your Own Latent: {A} New Approach to Self-Supervised
  Learning.
\newblock \emph{CoRR} abs/2006.07733.
\newblock \urlprefix\url{https://arxiv.org/abs/2006.07733}.

\bibitem[{He et~al.(2020)He, Fan, Wu, Xie, and Girshick}]{he2020momentum}
He, K.; Fan, H.; Wu, Y.; Xie, S.; and Girshick, R. 2020.
\newblock Momentum Contrast for Unsupervised Visual Representation Learning.

\bibitem[{Hjelm et~al.(2019)Hjelm, Fedorov, Lavoie-Marchildon, Grewal, Bachman,
  Trischler, and Bengio}]{hjelm2018learning}
Hjelm, R.~D.; Fedorov, A.; Lavoie-Marchildon, S.; Grewal, K.; Bachman, P.;
  Trischler, A.; and Bengio, Y. 2019.
\newblock Learning deep representations by mutual information estimation and
  maximization.
\newblock In \emph{International Conference on Learning Representations}.
\newblock \urlprefix\url{https://openreview.net/forum?id=Bklr3j0cKX}.

\bibitem[{Kohonen(1998)}]{KOHONEN19981}
Kohonen, T. 1998.
\newblock The self-organizing map.
\newblock \emph{Neurocomputing} 21(1): 1--6.
\newblock ISSN 0925-2312.
\newblock \doi{https://doi.org/10.1016/S0925-2312(98)00030-7}.
\newblock
  \urlprefix\url{https://www.sciencedirect.com/science/article/pii/S0925231298000307}.

\bibitem[{Li et~al.(2020)Li, Hu, Liu, Peng, Zhou, and Peng}]{li2020contrastive}
Li, Y.; Hu, P.; Liu, Z.; Peng, D.; Zhou, J.~T.; and Peng, X. 2020.
\newblock Contrastive Clustering.

\bibitem[{Sundareswaran et~al.(2021)Sundareswaran, Herrera-Gerena, Just, and
  Jannesari}]{sundareswaran2021cluster}
Sundareswaran, R.; Herrera-Gerena, J.; Just, J.; and Jannesari, A. 2021.
\newblock Cluster Analysis with Deep Embeddings and Contrastive Learning.

\bibitem[{Tan and Le(2020)}]{tan2020efficientnet}
Tan, M.; and Le, Q.~V. 2020.
\newblock EfficientNet: Rethinking Model Scaling for Convolutional Neural
  Networks.

\bibitem[{van~den Oord, Li, and Vinyals(2019)}]{oord2019representation}
van~den Oord, A.; Li, Y.; and Vinyals, O. 2019.
\newblock Representation Learning with Contrastive Predictive Coding.

\bibitem[{Wu et~al.(2018)Wu, Xiong, Yu, and Lin}]{wu2018unsupervised}
Wu, Z.; Xiong, Y.; Yu, S.; and Lin, D. 2018.
\newblock Unsupervised Feature Learning via Non-Parametric Instance-level
  Discrimination.

\bibitem[{Zitko(1994)}]{ZITKO1994718}
Zitko, V. 1994.
\newblock Principal component analysis in the evaluation of environmental data.
\newblock \emph{Marine Pollution Bulletin} 28(12): 718--722.
\newblock ISSN 0025-326X.
\newblock \doi{https://doi.org/10.1016/0025-326X(94)90329-8}.
\newblock
  \urlprefix\url{https://www.sciencedirect.com/science/article/pii/0025326X94903298}.

\end{thebibliography}

\end{document}